# Laplace's Method Approximations for Probabilistic Inference in Belief Networks with Continuous Variables

**Adriano Azevedo-Filho**[*]
*adriano@leland.stanford.edu*

**Ross D. Shachter**
*shachter@camis.stanford.edu*

Department of Engineering-Economic Systems
Stanford University, CA 94305-4025

## Abstract

Laplace's method, a family of asymptotic methods used to approximate integrals, is presented as a potential candidate for the tool box of techniques used for knowledge acquisition and probabilistic inference in belief networks with continuous variables. This technique approximates posterior moments and marginal posterior distributions with reasonable accuracy [errors are $O(n^{-2})$ for posterior means] in many interesting cases. The method also seems promising for computing approximations for Bayes factors for use in the context of model selection, model uncertainty and mixtures of *pdfs*. The limitations, regularity conditions and computational difficulties for the implementation of Laplace's method are comparable to those associated with the methods of *maximum likelihood* and *posterior mode analysis*.

## 1 Introduction

This paper provides an introduction to Laplace's method, a family of asymptotic techniques used to approximating integrals. It argues that this method or family of methods might have a place in the tool box of available techniques for dealing with inference problems in belief networks using the continuous variable framework. Laplace's method seems to accurately approximate posterior moments and marginal posterior distributions in belief nets with continuous variables, in many interesting situations. It also seems useful in the context of modeling and classification when used to approximate Bayes factors and posterior distributions of alternative models.

The ideas behind Laplace's method are relatively old and can be traced back, at least, to the developments presented by Laplace in one of his first major articles [Laplace 1774, p.366–367]. Since then they have been successfully applied in many disciplines. Some improvements in the implementation of Laplace's method introduced during the 80s induced a renewed interest in this technique, especially in the field of statistics, and will be discussed in the next section.

Initially, Section 2 presents an introduction to Laplace's method and its use in approximations for posterior moments and marginal posterior distributions. It also includes a discussion on the use of Laplace's method in approximations to Bayes factors and posterior *pdfs* of alternative models in general and in the particular case of mixtures of distributions. Section 3 discusses some implementation issues and limitations usually associated with the method. Section 4 illustrates Laplace's method with an inference problem from the medical domain. Finally, Section 5 presents some conclusions and recommendations.

## 2 Laplace's Method and Approximations for Probabilistic Inference

The approaches for probabilistic inference in belief networks with continuous variable usually consider techniques like: (a) analytical methods using conjugate priors [Berger 1985]; (b) linear iterative approximations using transformed variables and Gaussian influence diagrams results [Shachter 1990]; (c) numerical integration methods [Naylor and Smith 1982]; (d) simulation and importance sampling [Geweke 1989, Eddy et al. 1992]; (e) posterior mode analysis and maximum likelihood estimates [Eddy et al. 1992]; (f) discrete approximations [Miller and Rice 1983]; (g) mixtures of probability distributions [Poland and Shachter 1993, West 1993]; and (h) moment matching methods [Smith 1993]. Frequently these approaches can be combined and all of them can be useful in the context of specific problems.

---

[*]Also with the University of São Paulo, Brazil.

These techniques are, to varying degrees, well established in the belief networks/artificial intelligence/decision analysis literature and have been useful for dealing with many problems associated with learning and probabilistic inference. As an example, some of them are major elements in the confidence profile method [Eddy et al. 1992], a current belief network approach to deal with the synthesis of probabilistic evidence from scientific studies (meta-analysis) aimed primarily for the medical domain.

There is, however, another family of techniques based on asymptotic approximations of integrals and usually associated with the denomination *Laplace's method* or *Laplace's approximation* that also seems promising for dealing with some interesting instances from the same class of problems. This approach is introduced in the following paragraphs with some historical remarks.

Laplace's method and similar developments are inspired by the ingenious procedure used by Laplace in one of his first major papers [Laplace 1774, p. 366–367] to evaluate a particular instance of the integral

$$I_n = \int_a^b b(t) \exp(-n\, r(t))\, dt, \quad (1)$$

where $n$ is a large positive number, $r(t)$ is continuous, unimodal, and twice differentiable, having a minimum at $\hat{t} \in (a, b)$ and $b(t)$ is continuous, differentiable, and nonzero at $\hat{t}$. The general idea behind the solution arises from the recognition that with a "large" $n$ the most important contribution to the value of the integral comes from the region close to $\hat{t}$, the minimum of $r(t)$. An intuitive argument for the approximation is presented in sequence. First, the Taylor series for $r(t)$ and $b(t)$ is expanded at $\hat{t}$, leading to

$$I_n \approx \int_a^b (b(\hat{t}) + b'(\hat{t})(t - \hat{t})) \cdot e^{-n(r(\hat{t}) + r'(\hat{t})(t-\hat{t}) + \frac{1}{2}r''(\hat{t})(t-\hat{t})^2)}\, dt, \quad (2)$$

then, recognizing that $r'(\hat{t}) = 0$, and keeping only leading terms,

$$I_n \approx b(\hat{t}) e^{-n\, r(\hat{t})} \int_a^b e^{-\frac{n}{2} r''(\hat{t})(t-\hat{t})^2}\, dt; \quad (3)$$

finally, the limits of the integral are heuristically extended to $\infty$ and an unnormalized Gaussian *pdf* is recognized and integrated[1]. From this result follows the usual formula for Laplace's approximation in one dimension:

$$I_n \approx b(\hat{t}) \left(\frac{2\pi}{n}\right)^{\frac{1}{2}} \left(\frac{1}{r''(\hat{t})}\right)^{\frac{1}{2}} e^{-n\, r(\hat{t})}. \quad (4)$$

The approximation of $I_n$ by equation (4) is a standard result in the literature on asymptotic techniques shown to have an error term that is $O(n^{-1})$. Rigorous proofs for the approximation and behavior of errors, as well as lengthy discussions on assumptions and extensions are found in references like [De Bruijn 1961, p.36–39], [Wong 1989, p.55–66] and [Barndorff-Nielsen and Cox 1989, p.58–68]. Important extensions of these results include the cases: (a) $r(t)$ also dependent on $n$ [Barndorff-Nielsen and Cox 1989, p.61]; (b) $\hat{t} \in [a, b]$; and (c) multimodal functions.

Similar results also hold for more than one dimension. If now $\mathbf{t} \in \Re^m$ and the integration is performed over a $m$ dimensional domain, Laplace's approximation for the integral in equation (1) is just the $m$ dimensional extension of equation (4) [Wong 1989, p.495–500]:

$$I_n \approx b(\hat{\mathbf{t}}) \left(\frac{2\pi}{n}\right)^{\frac{m}{2}} (\det \Sigma_{\hat{\mathbf{t}}})^{\frac{1}{2}} e^{-n\, r(\hat{\mathbf{t}})}, \quad (5)$$

where $\hat{\mathbf{t}}$ is a point in $\Re^m$ where $\nabla r(\hat{\mathbf{t}})$, the gradient of $r(\mathbf{t})$ at $\hat{\mathbf{t}}$, is zero, and $\Sigma_{\hat{\mathbf{t}}}$, the inverse of the Hessian of $r(\cdot)$ evaluated at $\hat{\mathbf{t}}$, is assumed positive definite (meaning that $r(\mathbf{t})$ has a strict minimum at $\hat{\mathbf{t}}$). The general assumptions for this result are not unreasonable: unimodality, continuity on $b(\mathbf{t})$ and continuity on the second order derivatives of $r(\mathbf{t})$ in the neighborhood of $\hat{\mathbf{t}}$ [Wong 1989, p.498].

These results have had important applications in statistics. Laplace himself developed and used the procedures in a proof associated with what seems to be the first bayesian developments [Laplace 1774] after Bayes. Nevertheless, only during the last decades have these results started to be considered more seriously by statisticians in the context of practical applications [Mosteller and Wallace 1964, Johnson 1970, Lindley 1980, Leonard 1982].

A clever development presented by Tierney and Kadane [1986], later followed by a sequence of papers that also included the author R. E. Kass, inspired renewed interest on Laplace's method in recent years.Tierney and Kadane [1986] argued in favor of a specific implementation of Laplace's approximation, called by them *fully exponential*, that produces results that are more accurate than the ones obtained using other approaches. Instead of errors that are typically $O(n^{-1})$ for the conventional Laplace's approximation, they found that with their approach the errors are $O(n^{-2})$ due to the cancellation of $O(n^{-1})$ error terms. In previous developments [Mosteller and Wallace 1964, Johnson 1970, Lindley 1980] the same accuracy is achieved only when terms including derivatives of higher order are not dropped from the approximations, leading to formulas that are often difficult to apply in practice. In addition to that, Tierney and Kadane [1986] presented procedures to compute marginal posterior distributions, extending some ideas originally suggested by Leonard [1982]. In a sequence of papers [Kass et al. 1988, Tierney et al. 1989a; 1989b, Kass et al. 1990] the original intuitive developments presented by Tierney and Kadane [1986] were augmented by more formal derivations of the results.

---

[1] A sufficiently large $n$ can make the contribution from the region on the domain that does not include $[\hat{t} - \epsilon, \hat{t} + \epsilon]$ to the value of the integral arbitrarily small for any fixed small $\epsilon$. Similar argument can be used to eliminated the contribution of the term that includes $b'(\hat{t})(t - \hat{t})$.

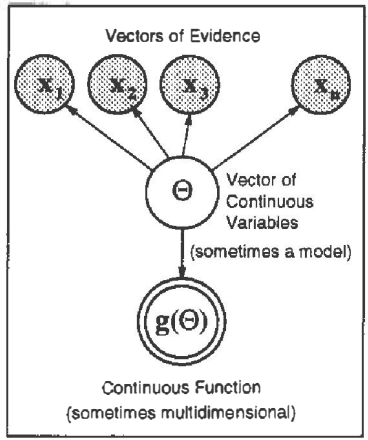

Figure 1: Belief net for the general problem

Laplace's method results have indeed a more general interpretation that can be extended to the context of belief networks and influence diagrams. The class of problems considered in the next sections is described by the belief net depicted on Figure 1. An important issue in this case might be the implementation of procedures to perform probabilistic inference on a function $g(\Theta)$ of a vector $\Theta = \{\theta_1, \theta_2, \ldots, \theta_m\}$ of continuous variables conditional on evidence represented by $\mathbf{X} = \{\mathbf{x}_1, \mathbf{x}_2, \cdots, \mathbf{x}_n\}$. This requires, generally speaking, constructing an arc from $\mathbf{X}$ to $g(\Theta)$. Another important issue might be the selection of models itself conditional on the evidence and prior beliefs. In both cases, when the conditions for applicability hold, Laplace's method seems to be a valuable technique.

Each instance of the evidence relates to $\Theta$ usually through a likelihood function that considers at least some of the elements from $\Theta$ as parameters of continuous probability density functions. This representation does not characterize the relationship among the elements of $\Theta$ that can be quite complex in some problems. The elements of $\Theta$ will frequently be called *parameters* because at least some of them will (possibly) be parameters of a specific probability density function.

Important results associated with Laplace's method are examined in the next subsections.

## 2.1 Approximations for posterior moments

To start the developments consider the definition of $E(g(\Theta)|\mathbf{X})$ in terms of the likelihood function and prior *pdf* on $\Theta$, a random vector:

$$E(g(\Theta)|\mathbf{X}) = \frac{\int_{\Omega_\Theta} g(\Theta) L(\mathbf{X}|\Theta) \pi(\Theta) \, d\Theta}{\int_{\Omega_\Theta} L(\mathbf{X}|\Theta) \pi(\Theta) \, d\Theta}. \qquad (6)$$

The first step in deriving Laplace's approximation to equation (6) involves the restatement of $g(\Theta)L(\mathbf{X}|\Theta)\pi(\Theta)$ and $L(\mathbf{X}|\Theta)\pi(\Theta)$ in the forms $b_1(\Theta)\exp(-nr_1(\Theta))$ and $b_2(\Theta)\exp(-nr_2(\Theta))$ so that the result in equation (5) can be easily applied. There are, indeed, infinite choices for the functions $b_i$ and $r_i$ in this case. The choice selected by Tierney and Kadane [1986], called *fully exponential*, leads to improved accuracy and considers:

$$b_1(\Theta) = b_2(\Theta) = 1 \qquad (7)$$

$$r_1(\Theta) = -n^{-1} \log(g(\Theta) L(\mathbf{X}|\Theta) \pi(\Theta)) \qquad (8)$$

$$r_2(\Theta) = -n^{-1} \log(L(\mathbf{X}|\Theta) \pi(\Theta)). \qquad (9)$$

Using these choices Tierney and Kadane [1986] argued that an approximation for $E(g(\Theta)|\mathbf{X})$ with $O(n^{-2})$ error terms can be obtained from the quotient of Laplace's approximation by equation (5) of each of the integrals on the numerator and denominator of equation (6). The improved result is derived from the convenient cancellation of $O(n^{-1})$ errors terms from each approximation. The expression for this approximation for $E(g(\Theta)|\mathbf{X})$ is

$$\hat{E}_n(g(\Theta)|\mathbf{X}) = \left(\frac{\det \Sigma_{\hat{\Theta}_1}}{\det \Sigma_{\hat{\Theta}_2}}\right)^{\frac{1}{2}} e^{-n(r_1(\hat{\Theta}_1) - r_2(\hat{\Theta}_2))}, \qquad (10)$$

where $\hat{\Theta}_1$ and $\hat{\Theta}_2$ are, respectively, the minimizers for $r_1(\Theta)$ and $r_2(\Theta)$ and $\Sigma_{\hat{\Theta}_i}$ is the inverse of the Hessian of $r_i(\Theta)$ evaluated at $\hat{\Theta}_i$. In this case

$$E(g(\Theta)|\mathbf{X}) = \hat{E}_n(g(\Theta)|\mathbf{X})(1 + O(n^{-2})). \qquad (11)$$

This result depends on a set of conditions more specific than those required for the conventional application of Laplace's Method that is referred as *Laplace regularity*. The conditions for *Laplace regularity* require, in addition to other aspects, that the integrals in equation (10) must exist and be finite, the determinant of the Hessians be bounded away from zero at the optimizers, and that the log-likelihood be differentiable (from first to sixth order) on the parameters and all the partial derivatives be bounded in the neighborhood of the optimizers. These conditions imply, under mild assumptions, asymptotic normality of the posterior. For formal proofs for the asymptotic results and extensive technical details on *Laplace regularity* see Kass et al. [1990].

The application of equation (10) to approximate $\text{Var}(g(\Theta)|\mathbf{X})$ and $\text{Cov}(g_1(\Theta), g_2(\Theta)|\mathbf{X})$ using the expressions (omitting $\Theta$):

$$\hat{\text{Var}}_n(g|\mathbf{X}) = \hat{E}_n(g^2|\mathbf{X}) - \hat{E}_n^2(g|\mathbf{X}) \quad (12)$$

$$\begin{aligned}\hat{\text{Cov}}_n(g_1, g_2|\mathbf{X}) &= \hat{E}_n(g_1 g_2|\mathbf{X}) \\ &\quad - \hat{E}_n(g_1|\mathbf{X})\hat{E}_n(g_2|\mathbf{X})\end{aligned} \quad (13)$$

also leads to accurate results [Tierney and Kadane 1986] as:

$$\text{Var}(g|\mathbf{X}) = \hat{\text{Var}}_n(g|\mathbf{X})(1 + O(n^{-2})) \quad (14)$$

and

$$\text{Cov}(g_1, g_2|\mathbf{X}) = \hat{\text{Cov}}_n(g_1, g_2|\mathbf{X}) + O(n^{-3}). \quad (15)$$

An aspect of using equation (8) that might seem restrictive in certain cases is the implied assumption that $g(\Theta)$ must be a nonnegative function (as $L(\mathbf{X}|\Theta)\pi(\Theta)$ is always nonnegative). This case can be addressed by at least two alternative approaches. The first one considers setting $h(\Theta, s) = \exp(s g(\Theta))$ (that is always nonnegative), computing Laplace's approximation for $E(h(\Theta))$, the moment generating function (*mgf*) for $g(\Theta)$, fixing s at a convenient value where the *mgf* is defined, and then using the approximation $\hat{E}(g(\Theta)) = \frac{\partial \hat{E}(h(\Theta,s))}{\partial s}|_{s=0}$ (the definition of expectation from a *mgf* of a random variable). The second alternative consider setting $h(\Theta) = g(\Theta) + c$, where $c$ is a large positive value, computing Laplace's approximation for $E(h(\Theta))$ and using the approximation $\hat{E}(g(\Theta)) = \hat{E}(h(\Theta)) - c$. Both alternatives are shown [Tierney et al. 1989b] to be equivalent when $c \to \infty$, having absolute approximation errors that are $O(n^{-2})$.

An example is presented in sequence to illustrate the application of these results.

**Example 1 (Beta posterior):** *Experimental results showed that a coin flipped n times produced p heads and q tails. Let $\theta$ be probability of heads and assume that our prior knowledge on $\theta$ is represented by a Beta(a,b) pdf. Suppose that we want to compute an approximation for the posterior expected value of $\theta$ using Laplace's method.*

*In this case $g(\theta) = \theta$ and $L(\mathbf{X}|\theta)\pi(\theta) = c\,\theta^{p+a-1}(1-\theta)^{q+b-1}$ (c is some constant). The minimizers of $r_1(\theta)$ and $r_2(\theta)$, expressions defined in equation (8) and*

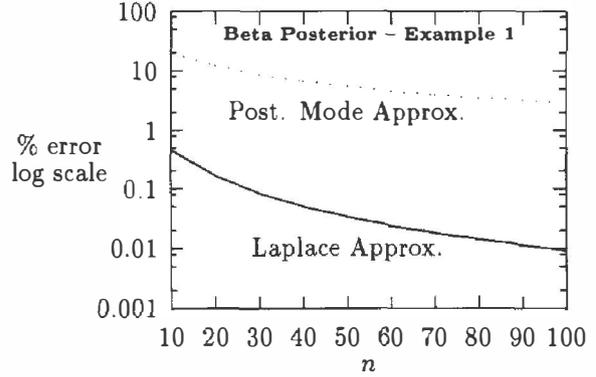

Figure 2: Errors in Approximations for $E(\theta|\mathbf{X})$.

*equation (9), can be easily computed and are, respectively, $\hat{\theta}_1 = \frac{p+a}{p+q+a+b-1}$ and $\hat{\theta}_2 = \frac{p+a-1}{p+q+a+b-2}$. Then, making the substitutions in equation (10), recalling that in this unidimensional example $\det \Sigma_i(\hat{\theta}_i)$ is just the inverse of the second derivative of $r_i(\theta)$ evaluated at the appropriate minimizer and letting $s = (p + a)$ and $r = (q + b)$, it follows that*

$$\hat{E}(\theta|\mathbf{X}) = \left(\frac{s^{2s+1}(s+r-2)^{2s+2r-1}}{(s-1)^{2s-1}(s+r-1)^{2s+2r+1}}\right)^{\frac{1}{2}}.$$

In this example, the error of Laplace's approximation can be easily examined as the analytical expression for $E(\theta|\mathbf{X})$ can be computed. In this case the posterior for $\theta$ is a $Beta(p+a, q+b)$ pdf so $E(\theta|\mathbf{X}) = \frac{p+a}{p+q+a+b}$ and $\text{Mode}(\theta|\mathbf{X}) = \frac{p+a-1}{p+q+a+b-2}$. The asymptotic behavior of the error is illustrated in Figure 2 in a situation where the $n = 10k$ is the number of flips, $p = 2k$ is the number of observed heads, $q = 8k$ is the number of tails and $a = b = 1$ are the parameters of the prior knowledge on $\theta$. As $k$ increases from 1 to 10, and $n$ varies from 10 to 100, the relative error from Laplace's approximation decreases with $n$, in a way that seems consistent with the expected asymptotic behavior. The same figure presents, for comparison, the behavior of the relative error from an approximation to the posterior mean using the posterior mode.

### 2.2 Approximations for Marginal Probability Distributions

Laplace's method can be also useful for approximations of marginal distributions. Two important cases are examined in this section: the approximation for a marginal posterior distribution and the general case of an approximation to a nonlinear function of parameters, conditional on the evidence $\mathbf{X}$.

Let $\Theta$ be partitioned into the subsets $\Theta_p$ and $\Theta_q$ ($q$ are the number of elements in each subset) and suppose that we are interested in computing the marginal pos-terior distribution for $\Theta_p$ in the light of the evidence $\mathbf{X}$ considering the same generic model described in Figure 1. Explicitly:

$$\pi(\Theta_p|\mathbf{X}) = c \cdot \int_{\Omega_{\Theta_q}} L(\mathbf{X}|\Theta_p, \Theta_q)\,\pi(\Theta_p, \Theta_q)\,d\Theta_q \quad (16)$$

for a constant $c$ that can be analytically defined by:

$$c = \int_{\Omega_{\Theta_p}, \Omega_{\Theta_q}} L(\mathbf{X}|\Theta_p, \Theta_q)\,\pi(\Theta_p, \Theta_q)\,d\Theta_p\,\Theta_q. \quad (17)$$

An approximation for equation (16) can be easily found for $\theta_p = k$ using two alternative approaches. The first approach considers the use of Laplace's method to approximate both the integral part in equation (16) and the constant $c$ defined in equation (17). In the second approach the constant $c$ is approximated by an external procedure, usually numerical integration, that is very effective in low dimensions (and frequently $p = 1$ or 2), according to Naylor and Smith [1982]. In this case, a set of approximated values for the integral in equation (16) is computed by repeated application of Laplace's method with $\Theta_p$ set to conveniently chosen values from its domain. The implementation of the second approach, using the fully exponential procedure, in equation (5) leads to the following approximation for the marginal pdf in equation (16) at $\Theta_p = k$:

$$\hat{\pi}_n(k|\mathbf{X}) = c \left(\det \Sigma_{k,\hat{\Theta}_q(k)}\right)^{\frac{1}{2}} e^{-n r(k, \hat{\Theta}_q(k))} \quad (18)$$

where $\hat{\Theta}_q(k)$ is the minimizer of $r(\Theta_p = k, \Theta_q)$ and $\Sigma_{k,\hat{\Theta}_q(k)}$ is the inverse of the Hessian of $r(\Theta_p = k, \Theta_q)$ evaluated at $\hat{\Theta}_q(k)$. In this case $c$ is computed by an external numerical procedure or even heuristically adjusted if what is needed is just a rough graphical characterization of the distribution.

The relative errors in the approximation of equation (16) by the first approach are $O(n^{-1})$. The second approach, $c$ computed numerically from the integration of the unnormalized marginal, leads to more accurate approximations, with errors that are $O(n^{-\frac{3}{2}})$. Another important aspect of these procedures is that they are surprisingly accurate in the approximation of the tail behavior of the distribution as the errors are $O(n^{-1})$ uniformly on all bounded neighborhoods of the mode [Kass et al. 1988]. In addition, it is also possible to show that this approximation has properties that are comparable to those obtained from saddlepoint approximations, a family of techniques that consider the application of Laplace's method in the domain of complex numbers (see Reid [1988] for more details on this technique).

The following example illustrates these results.

**Example 2 (Gaussian):** *Let $\mathbf{X}$ be a set of $n$ independent measurements from a gaussian population with parameters $\Theta = \{\theta_\mu, \theta_\sigma^2\} = \{\mu, \sigma^2\}$. Assume that the prior belief on the parameters is represented by $\pi(\theta_\mu, \theta_\sigma) = 1/\theta_\sigma$, $\theta_\sigma > 0$. Suppose that we want to compute the marginal posterior probability $\pi(\theta_\sigma|\mathbf{X})$ using Laplace's method.*

*In this example*

$$L(\mathbf{X}|\Theta)\pi(\Theta) \propto \theta_\sigma^{-(n+1)} e^{-\frac{(n-1)s^2 + n(\hat{\mu} - \theta_\mu)^2}{2\theta_\sigma^2}},$$

*where $\hat{\mu}$ and $s^2$, the mean and the variance of the measurements, summarize all the evidence from $\mathbf{X}$. Consider equation (18) to approximate the posterior marginal at $\theta_\sigma = k$, using equation (9) to define $r(\Theta)$. In this case $\hat{\theta}_\mu = \hat{\mu}$ is the minimizer of $r(k, \theta_\mu)$ and $\Sigma_{k,\theta_\mu} \propto k^2$. This leads to the following approximation for the posterior marginal:*

$$\pi(\theta_\sigma = k|\mathbf{X}) \propto k^{-n} e^{-\frac{(n-1)s^2}{2k^2}}$$

*This approximation is indeed proportional to the exact expression for the posterior marginal of $\theta_\sigma$ that is an inverted gamma. Laplace's approximation for the marginal posterior of $\theta_\mu$ is also proportional to the exact* pdf *that is a Student t in this case.*

A more general situation considers Laplace's approximation for a *pdf* of a nonlinear function $g(\Theta)$. The necessary conditions for using Laplace's approximation to this case require that, in the neighborhood of the mode of the density of $\Theta$, the gradient of $g(\Theta)$ is nonnull or the Jacobian of $g(\Theta)$ is full rank ($g(\Theta)$ can be a $m$ dimensional function). This condition ensures local invertibility of $g(\Theta)$ in the region close to the mode of the distribution (the distribution is assumed unimodal). If these conditions hold, an approximation to the density of $g(\theta)$ can be constructed using Laplace's Method, as showed by Tierney et al. [1989a]. A possible alternative for this approximation is

$$\hat{\pi}_n(g(\Theta) = k|\mathbf{X}) = c\,A\,e^{-n\,r(\hat{\Theta}(k))} \quad (19)$$

where

$$A = \left(\frac{\det \Sigma_{\hat{\Theta}(k)}}{\det(\nabla g(\hat{\Theta}(k))^T \Sigma_{\hat{\Theta}(k)} \nabla g(\hat{\Theta}(k)))}\right)^{\frac{1}{2}} \quad (20)$$

and

$$\hat{c} = (2\pi)^{-\frac{m}{2}} \left(\det \Sigma_{\hat{\Theta}}\right)^{-\frac{1}{2}} e^{n\,r(\hat{\Theta})} \quad (21)$$

$\hat{\Theta}$ and $\hat{\Theta}(k)$ are, respectively, the minimizers of $r(\Theta)$ and $r(\Theta)$ subject to the constraint $g(\Theta) = k$; $\nabla g(\hat{\Theta}(k))$ is the gradient (or Jacobian) of $g(\Theta)$ evaluated at $\hat{\Theta}(k)$ (a $p \times k$ matrix), with p being the number of elements in $\Theta$ and $m$ the dimensions of the function $g(\Theta)$, that is frequently 1; and, $\Sigma_{\hat{\Theta}(k)}$ is the inverse of the Hessian of $r(\Theta)$ evaluated at the appropriate minimizer.

The same considerations about the normalization constant discussed previously also apply to this case. The use of equation (21) as an approximation to the constant in equation (19) does not ensure that it will integrate to one. However, it can be a good approximation for many purposes associated with graphical characterization of the probability function. The errors in this approximation follow the same behavior as in the first situation analyzed in this subsection, as showed by Tierney et al. [1989a].

## 2.3 Approximations for Bayes Factors, Model Selection and Mixtures

In recent years there has been a renewed interest in the use of the *Bayes factors* as an important tool for testing hypothesis, selecting models, and dealing with model uncertainty. In a general context, for a set of evidence $\mathbf{X}$ and two alternative models or hypothesis $M_1$ and $M_2$ that include, respectively, the sets of continuous parameters $\Theta_1$ and $\Theta_2$, Bayes factor is defined by

$$b_{12} = \frac{\int_{\Omega_{\Theta_1}} L(\mathbf{X}|\Theta_1, M_1)\pi(\Theta_1|M_1)d\Theta_1}{\int_{\Omega_{\Theta_2}} L(\mathbf{X}|\Theta_2, M_2)\pi(\Theta_2|M_2)d\Theta_2} \quad (22)$$

so that

$$\frac{\pi(M_1|\mathbf{X})}{\pi(M_2|\mathbf{X})} = b_{12} \cdot \frac{\pi(M_1)}{\pi(M_2)} \quad (23)$$

Analytical solutions for Bayes factors are only available in specific cases. For more general situations, Bayes factors are computed by Monte Carlo simulation, numerical integration and approximation methods that include Laplace's method and some variants – see Kass and Raftery [1994] for an extensive overview that includes many applications.

A possible approximation for the Bayes factor using Laplace's method considers equation (5) with $b(\cdot) = 1$ and

$$r(\Theta_i|M_i) = -n^{-1}\log(L(\mathbf{X}|\Theta_i, M_i)\pi(\Theta_i|M_i)), \quad (24)$$

$i = 1, 2$, to approximate both the numerator and denominator of equation $(22)^2$. The expression for this approximation is

$$\hat{b}_{12} = (2\pi)^{\frac{m_1-m_2}{2}}\left(\frac{\det \Sigma_{\hat{\Theta}_1}}{\det \Sigma_{\hat{\Theta}_2}}\right)^{\frac{1}{2}} e^{-n(r(\hat{\Theta}_1)-r(\hat{\Theta}_2))} \quad (25)$$

---

[2]This case uses a modified version of equation (5) that is derived using $r(\cdot)$ defined directly as a function of $n$ so that the term on $n$ under $\pi$ does not appear in the expression.

where $m_i$ is the dimension of $\Theta_i$. It has an approximation error that is $O(n^{-1})$. A variant of this approach considers the maximum likelihood estimator for $\Theta_i$ instead of the minimizer of $r(\Theta)$ and the inverse of the expected information matrix instead of the inverse of the Hessian in equation (25). This alternative might be convenient when there is statistical software available that is able to perform these computations. In this case the error in the approximation is also $O(n^{-1})$ but if the prior is informative the result is likely to be less accurate than the result from Laplace's approximation [Kass and Raftery 1994].

The following example illustrates these results.

**Example 3 (Bayes factor):** *A group of $n$ subjects was randomly split in two subgroups that were, respectively, submitted to two alternative treatments, $T_1$ and $T_2$. The possible outcomes of the experiment for each subject are either 1 or 2. Let $n_{ij}$, $i = 1, 2$ and $j = 1, 2$, be the number of subjects that received treatment $i$ and presented outcome $j$ and $\theta_i$ be the probability of a subject that received treatment $i$ presented outcome 1. Suppose that we want to get some evidence about whether the treatments induced different $(M_1)$ or identical $(M_2)$ outcomes using the Bayes factor. Assume that the prior knowledge is represented by $\pi(\theta_1, \theta_2|M_1) = 1$ and $\pi(\theta|M_2) = 1$.*

*In this case the analytical expression for the Bayes factor is*

$$b_{12} = \frac{B(n_{11}+1, n_{12}+1)B(n_{21}+1, n_{22}+1)}{B(n_{11}+n_{21}+1, n_{12}+n_{22}+1)},$$

*where $B(\cdot, \cdot)$ is the beta function. Laplace's approximation can be easily found using equation (25):*

$$\hat{b}_{12} = \frac{(2\pi)^{\frac{1}{2}}n^{n+\frac{1}{2}}n_{11}^{n_{11}+\frac{1}{2}}n_{12}^{n_{12}+\frac{1}{2}}n_{21}^{n_{21}+\frac{1}{2}}n_{22}^{n_{22}+\frac{1}{2}}}{p^{p+\frac{3}{2}}q^{q+\frac{3}{2}}r^{r+\frac{1}{2}}s^{s+\frac{1}{2}}}$$

*where $p = (n_{11}+n_{12})$, $q = (n_{21}+n_{22})$, $r = (n_{11}+n_{21})$ and $s = (n_{12}+n_{22})$. The asymptotic behavior of the error in the approximation is presented in Figure 3 for $n_{11} = 3k$, $n_{12} = 2k$, $n_{21} = 4k$, $n_{22} = 1k$, $n = 10k$, with $k$ increasing from 1 to 10.*

In the context of model uncertainty, Bayes factors can be used to compute the posterior probability of possible models, given the evidence provided by $\mathbf{X}$, as a direct extension of equations (22) and (23):

$$\pi(M_i|\mathbf{X}) = \frac{\frac{\pi(M_i)}{\pi(M_j)} \cdot b_{ij}}{\sum_{k \in \Omega_M} \frac{\pi(M_k)}{\pi(M_j)} \cdot b_{kj}} \quad (26)$$

Laplace's approximation is derived by replacing the exact values of the Bayes factors in equation (26) by approximations using equation (25) with the appropriate indexes. An important issue in the use of these approximations is whether the conditions for *Laplace regularity* hold in a particular case to ensure the applicability of Laplace's method. This issue is examined in the next paragraphs in the context of models involving mixtures of distributions.

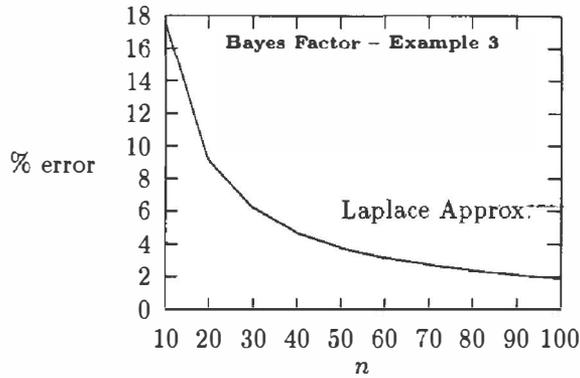

Figure 3: Errors in Approx. for the Bayes Factor

A promising area for Laplace's method, closely related to model uncertainty, is the approximation of posterior probabilities of the number of components in mixtures of distributions and group classification. This problem is a particular instance of equation (26) where each model relates to a certain discrete value for the number of components in the mixture. Laplace's method is applied using the same approximation suggested for equation (24) considering in this case $M_k = $ {number of components in the mixture is k} and

$$L(\mathbf{X}|\Omega_k)\pi(\Omega_k) = \left[\prod_{i=1}^{n}\sum_{j=1}^{k}\lambda_j P_j(\mathbf{x}_i|\Theta_j)\right]\pi(\Omega_k) \quad (27)$$

where $\Omega_k = \{\bigcup_{j=1}^{k}\Theta_j, \bigcup_{j=1}^{k}\lambda_j\}$, $\lambda_j \in [0,1]$ with $\Sigma_{j=1}^{k}\lambda_j = 1$, $\mathbf{X} = \{\mathbf{x}_1, \mathbf{x}_2, \cdots, \mathbf{x}_n\}$ and $P_j(\cdot)$, $j = 1, \ldots, k$, are *pdfs* (usually from the same parametric family).

Two aspects make this problem interesting in the context of Laplace's method. First, if the mixture is *identifiable* – and mixtures with *pdfs* from the same parametric family are identifiable in most cases – there is still an obvious problem of multimodality in equation (27) due to the possibility of label switching. This problem can be overcome with constraints to avoid label switching in the process of finding the minimizer for equation (24). Second, there is the concern about whether *Laplace regularity* holds in this case. Indeed it does hold in this case at least for a large class of *pdfs* that includes the exponential family as well as other important parametric families. For more details on Laplace's method in mixtures as well as proofs for the regularity conditions and some applications see Crawford [1994].

## 3 Implementation Issues and Limitations

The computational implementation of Laplace's method is relatively straightforward. It depends on the availability of the same computer routines that implement classical optimization procedures used by the methods of *maximum likelihood* and *posterior mode analysis*. The approximation of marginal distributions usually requires numerical integration procedures – if accurate approximations for the integration constant or probabilities of regions of the distribution are needed – as well as plotting procedures.

Typical applications of Laplace's method to the approximation of moments involve the computation of two minimizers for the expressions in equations (8) and (9), one of them being usually the posterior mode. Frequently the second minimizer is found with only a few steps of Newton's method (1 to 3 usually) when one of the minimizers is used as the starting point for the second optimization process. Similar procedure can be used in the case of marginals, considering in these case information from previous approximations.

This means that the computational effort needed to implement Laplace's method is only marginally greater than that needed for the *posterior mode analysis* or *method of maximum likelihood.*   One aspect that seems critical, however, is the availability of improved computational procedures to find numerical approximations to gradients and specially to Hessians if the analytical expressions are not available. This point was stressed by Naylor [1988] and also applies to other methods in some extension (when Newton's method is used for example).

Reparametrization is shown to be a useful practice in some cases to make the posterior distribution look closer to a Gaussian [Tierney et al. 1989b, Crawford 1994].

Other limitations are in general related to possible problems with multimodality in the distribution and the lack of practical diagnostic procedures to ensure *a priori* the asymptotic properties of the method. Even if the asymptotic behavior holds in a particular situation it does not mean that a particular approximation is accurate. In this case experience seems to be the best guide [Kass et al. 1988].

## 4 An Application to a Medical Inference Problem

This section illustrates the use of Laplace's method with a problem from the medical domain (Figure 4). The problem involves the use of tamoxifen and was previously modeled and analyzed elsewhere [Eddy et al. 1992, p.183–189]. Part of this previous study considered the posterior mode analysis to access the increase in the probability of one year survival without disease from alternative treatments, in the light of the evidence provided by two medical studies. This quantity is represented here by $\varepsilon$.

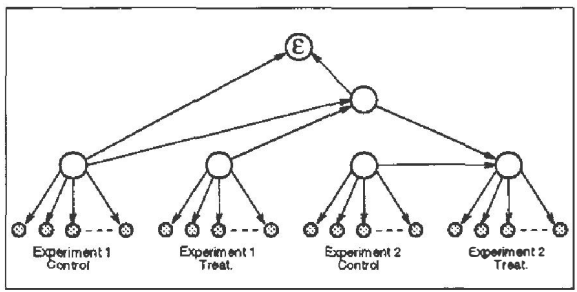

Figure 4: Belief net for the medical problem

Table 1: Errors in Approximations for $\hat{E}(\varepsilon|\mathbf{X})$.

| Method | $\hat{E}(\varepsilon|\mathbf{X})$ |
|---|---|
| Monte Carlo | 0.05694 |
|  | $(0.00054)^a$ |
| Laplace's method | 0.05686 |
|  | $(0.14\%)$ |
| Posterior Mode[b] | 0.05832 |
|  | $(2.42\%)^c$ |

[a]Standard deviation.
[b]Eddy et al. [1992] reported 0.059.
[c]compared to Monte Carlo.

To allow some comparison among alternative methods, an approximation for the posterior expected value of $\varepsilon$ was computed using a Monte Carlo procedure with importance sampling that considered $10^6$ samples. Then, an implementation of Laplace's method, involving numerical methods to compute gradients and Hessians, was used to find an approximation for the posterior expected value of $\varepsilon$ – the posterior mode of $\pi(\varepsilon|\mathbf{X})$ was found as an intermediate step in the computations. These results are presented in Table 1.

If the Monte Carlo approximation is arbitrarily taken as the "gold standard" for $\hat{E}(\varepsilon|X)$, the relative errors of Laplace's and posterior mode approximations are, respectively, 0.14% and 2.42%. A realistic benchmark for alternative techniques would certainly require more extensive study. Nonetheless, this example illustrates the application of Laplace's method to a realistic problem, even though the results don't change the conclusions from the previous analysis of that experiment done by Eddy et al. [1992].

An interesting extension for cases like this is the approximation of the marginal posterior *pdf* as a way to get extra insights into the problem. Laplace's approximation for $\pi(\varepsilon|\mathbf{X})$ is presented in Figure 5. It was computed using equation (18) with the constant $c$ estimated by numerical integration. The approximations were found for values of $\varepsilon$ on the interval $[-0.08, 0.20]$ considering points spaced 0.005 apart using Newton's method to compute the minimizers. Figure 5 also shows a Monte Carlo approximation for $\pi(\varepsilon|\mathbf{X})$ using $2 \times 10^4$ samples classified into 50 classes in the interval $[-0.08, 0.20]$. The points in the figure represent the frequency density of each class at the center of the class. The computations for Monte Carlo (including classification) took roughly 20 times longer than Laplace's method computations.

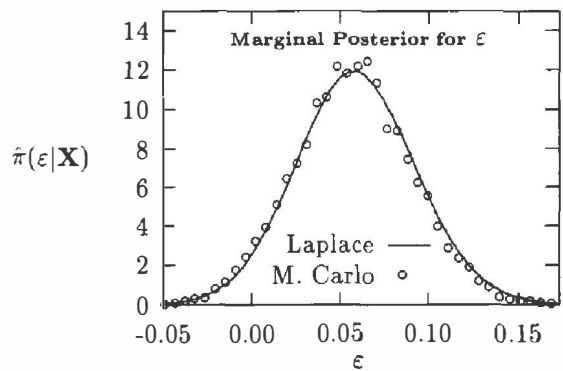

Figure 5: Laplace's Approximation for $\pi(\varepsilon|\mathbf{X})$

## 5 Final Considerations

Laplace's method can be viewed as an interesting extension of methods like posterior mode analysis and maximum likelihood as it uses similar implementation procedures, shares with them some of the same problems, but extends their functionality. Laplace's method directly computes an approximation for moments that seems reasonably accurate for many uses, possibly avoiding Monte Carlo methods in some cases. This feature might be useful in problems where the moment of the quantity of interest has special meaning (e.g. expected utility in the context of influence diagrams in decision analysis). Another useful extension is the approximation for marginal distributions so that they can be easily plotted or numerically integrated to produce probabilistic statements about the quantity of interest. An area of potential interest for investigation seems to be the combination of Laplace's method with Monte Carlo methods (e.g. in approximations of the posterior using mixtures of parametric distributions and in the formulation of importance sampling distributions).

### Acknowlegements

A. Azevedo-Filho's research was supported by a grant from FAPESP, Brazil. We would like to thank three anonymous referees for their helpful suggestions.